\documentclass[runningheads]{llncs}
\usepackage[T1]{fontenc}
\usepackage{tikz}
\usepackage{float}
\usepackage{siunitx}
\usepackage{multirow}
\usepackage{graphicx}
\usepackage{array}
\usepackage{booktabs}
\usepackage{amsmath}
\usepackage{hyperref}
\usepackage{csquotes}
\usepackage{cleveref}
\usepackage{mwe}

\usepackage[acronym]{glossaries}
\usepackage{textcomp}
\usepackage{amsfonts}
\usepackage{bm}
\usepackage{amssymb}
\usepackage{pifont}
\newcommand{\cmark}{\ding{51}} 
\newcommand{\xmark}{\ding{55}} 

\newcommand{\std}[1]{\textcolor{gray}{\scriptsize\textpm\ #1}}

\glsdisablehyper
\newacronym{mtsad}{MTSAD}{Multivariate Time Series Anomaly Detection}
\newacronym{ot}{OT}{Operational Technology}
\newacronym{ics}{ICS}{Industrial Control System}
\newacronym{mts}{MTS}{Multivariate Time Series}
\newacronym{ad}{AD}{Anomaly Detection}
\newacronym{aucpr}{AUC-PR}{Area Under the Precision-Recall Curve}
\newacronym{vuspr}{VUS-PR}{Volume Under the Precision-Recall Surface}
\newacronym{dl}{DL}{Deep Learning}
\newacronym{cl}{CL}{Centralized Learning}
\newacronym{fl}{FL}{Federated Learning}
\newacronym{hfl}{H-FL}{Hierarchical Federated Learning}
\newacronym{dof}{DoF}{Degrees of Freedom}
\newacronym{cps}{CPS}{Cyber-Physical System}
\newacronym{asd}{ASD}{Application Server Dataset}
\newacronym{qappd}{QAPPD}{Quanser Aero 2 Pick-and-Place Datase}
\newacronym{smd}{SMD}{Server Machine Dataset}
\newacronym{msl}{MSL}{Mars Science Laboratory}
\newacronym{smap}{SMAP}{Soil Moisture Active Passive}
\newacronym{swat}{SWaT}{Secure Water Treatment}
\newacronym{wadi}{WaDi}{Water Distribution}
\newacronym{psm}{PSM}{Pooled Server Metrics}

\begin{document}

    \title{Federated Learning for Multivariate Time Series Anomaly Detection in Industrial Automation}

    \titlerunning{\acrshort{fl} for \acrshort{mtsad} in Industrial Automation}


    \author{Khayyam Nosrati\inst{1} \and
    Martin Uray\inst{1,2} \and
    Saverio Messineo\inst{1} \and
    Olaf Sassnick \inst{1} \and
    Stefan Huber\inst{1}}


    \authorrunning{K. Nosrati et al.}

    \institute{
        Josef Ressel Centre for Intelligent and Secure Industrial Automation,\\
        Salzburg University of Applied Sciences, Austria\\
        \and
        Department of Artificial Intelligence and Human Interfaces,\\
        Paris Lodron University of Salzburg, Austria
    }

    \maketitle              

    \begin{abstract}
        Federated learning (\acrshort{fl}) has broadened the horizon for multivariate time series anomaly
        detection (\acrshort{mtsad}).
        However, benchmarking such anomaly detection methods within~\acrshort{fl} paradigm
        poses data-centric challenges. The existing datasets do not counteract these challenges since they do not
        simultaneously provide sufficient scale, accurate labels, and freedom from common flaws.
        In addition, the role of cyclic process behavior, which is common in discrete industrial automation,
        remains underexplored for \acrshort{mtsad} for the current state of research.
        This paper aims to shed more light on the literature and address these gaps by introducing a dataset designed
        with cyclic dynamics arising from the repetitive nature of discrete automation processes 
        and evaluates selected \acrshort{mtsad} methods on both the proposed dataset and a public benchmark dataset.
    \end{abstract}

    \section{Introduction}
    \begingroup
    \renewcommand\thefootnote{}\footnotetext{Preprint. Accepted at the DEXA International Workshop on
    Optimisation of Industrial Production with AI Algorithms 2026 (DEXA AI4IP 2026).\\
    Corresponding Author: M.~Uray (\href{mailto:martin.uray@fh-salzburg.ac.at}{martin.uray@fh-salzburg.ac.at}).}
    \endgroup

    \gls{ot} systems~\cite{2023nist}, such as \gls{ics}, are distributed, embedded cyber-physical infrastructures designed for
    real-time monitoring and control of industrial processes, and are the backbone
    of critical industrial sectors.
    The advent of intelligent technologies has enhanced the capabilities of these systems to better leverage
    data and foster a new era of automated and intelligent computing.
    One of the most prominent applications of AI in \acrshort{ot} environments is anomaly
    detection; and more particularly, anomaly detection on \gls{mts} data acquired through the observation of dynamic processes.

    Recent research has introduced advanced \gls{dl} techniques for anomaly detection, demonstrating high potential.
    \acrshort{dl} scales with input data volume in various tasks~\cite{sarker2021deep,hestness2017deep},
    which is one of the core motivation to centralizing data and learning an effective machine learning
    model on a central instance. This paradigm is known as \gls{cl} which typically handles
    massive computation and enormous storage demands on cloud servers.

    Alternatively, \acrlong{fl}, and \gls{hfl} provide distributed computation and the necessary communication topology to
    address the challenges of risen by \acrshort{cl}.

    In this work, we investigate the application of \acrshort{mtsad} in an unsupervised setting within the paradigms of \acrshort{cl}, \acrshort{fl}, and \acrshort{hfl}.
    We evaluate performance trade-offs on datasets with \emph{different} characters, while we primarily focus on
    horizontal federated learning and centralized learning setups.
    In horizontal \acrshort{fl} local datasets at edge nodes have identical and shared feature space, while sample space
    varies. Therefore, this study assumes homogeneous neural network architectures across edge nodes.

    \paragraph{Contributions.}%
    In this work, we identified the following key research gaps: (i) lack of comparative studies on federated and hierarchical
    federated \acrshort{mtsad}, (ii) absence of \acrshort{fl} repetitive \acrshort{mtsad}-specific datasets out of industrial
    production cycles.
    In addition, the lack of a unified definition of industrial anomaly, combined
    with the use of multiple, sometimes flawed evaluation metrics, prevents the
    development of consistent and robust assessment practices to advance industry
    and automation. To address these gaps, we make the following contributions:
    \begin{itemize}
        \item Introduce a new suite of ten repetitive multivariate time series anomaly detection datasets.
        \item Provide comparative insights on the five \acrshort{mtsad} methods that curated by a principled
              approach---based on problem-solving specificity, architectural suitability for \acrshort{mtsad},
              and integrability to available \acrshort{fl} software frameworks while conserving diversity of
              architectures---across centralized, federated, and hierarchical-federated learning paradigms on repetitive
        and non-repetitive datasets.
    \end{itemize}

    \section{Related Work}

    \acrlong{mtsad} has long been approached in localized learning environments, where machine learning is performed
    on a single device using a single dataset.
    Existing \acrshort{mtsad} approaches can broadly be categorized into statistical, forecasting-based~\cite{xu2021anomaly},
    and reconstruction-based~\cite{10.14778/3514061.3514067} methods.
    More recently, \acrshort{mtsad} is approached by federated learning such as FATRAF~\cite{TRUONG2022103692} which proposes a lightweight
    architecture designed for communication efficiency investigated in industrial environments, while~\cite{10.1155/2022/2913293}
    are studied in non-industrial applications. Different aggregation strategies are utilized by such methods:
    FATRAF and~\cite{HUONG2021103509} adapt FedAvg, FedUAD~\cite{10064694} employs the FedCC, and~\cite{10.1155/2022/2913293}
    integrates FLTrELM.
    Other advancements involve asynchronous communication protocol~\cite{raeiszadeh2025asynchronous}.
    Hierarchical federated learning has also been considered for \acrshort{mtsad}, as demonstrated by
    HFL-ADS~\cite{s24175492} and HFed-IDS~\cite{electronics11162627}, which apply hierarchical variants of
    FedAvg incorporating intermediate aggregation levels.
    Despite these developments, benchmarking studies comparing federated and hierarchical
    federated learning approaches for \acrshort{mtsad} remain largely absent.

    Several highly cited public benchmarks are repeatedly used for evaluating \acrshort{mtsad},
    either as single datasets or as collections of similar datasets~\cite{10.1145/3292500.3330672,10.14778/3476249.3476307,10.1145/3219819.3219845,10.1145/3447548.3467174}.
    \gls{smd} is a collection of $28$ datasets which represent a 5-week-long data
    collected from a large internet company.
    Exathlon is constituted by recording the repeated executions of 10
    different Spark streaming applications on a 4-node cluster.
    NASA’s \gls{msl} and \gls{smap} datasets contain multivariate time series
    where one variable representing a sensor measurement and the remaining variables represent one-hot encoded telemetry
    commands.
    The observations in \acrshort{smd}, \acrshort{msl}, \acrshort{smap} are all equally-spaced one minute apart,
    while in Exathlon are one second apart.
    Additional benchmarks such as \gls{swat}, \gls{wadi}, and \gls{psm}
    are commonly employed as single datasets for \acrshort{mtsad}.
    In federated settings, these datasets are typically adapted through partitioning strategies,
    where subsets of data are assigned to multiple edge nodes.
    However, prior analyses have highlighted that these suffer from inherent limitations or flaws~\cite{9835419,10.1145/3447548.3467075,Wagner2023TimeSeADBD}.
    Besides, these datasets lack explicit cyclic properties.

    In conclusion, there is scarcity of datasets suitable for \acrshort{mtsad} that capture discrete,
    repetitive, and state-driven characteristics of industrial production cycles, including pick-and-place operations,
    injection molding, assembly, and in-line component testing. Furthermore, current literature rarely investigates
    these scenarios through joint benchmarking of federated and hierarchical federated learning approaches.
    These shortcomings collectively motivate the focus of this work.

    \section{Preliminaries}\label{sec:preliminaries}

    \subsection{Federated Learning}\label{subsec:federated-learning}
    In this work, centralized, federated, and hierarchical federated learning settings follow a centralized communication topology.
    The choice of \acrshort{fl} aggregation strategy is \textit{FedAvg}~\cite{mcmahan2017communication}
    for its simplicity of implementation, interpretability and strong performance reported in the literature.
    For \acrshort{hfl} aggregation, HierFAVG~\cite{liu2020client}---an extended version of FedAvg that has theoretical
    convergence guarantees---is adopted. The default optimizers of the chosen methods are set to SGD for \acrshort{fl}
    and \acrshort{hfl} simulations, because FedAvg and HierFAVG use SGD as their local optimizers.

    \subsection{Evaluation Metrics}\label{subsec:prel-eval-metrics}
    This work adopts a centralized evaluation protocol that considers the union
    of all available windowed test datasets and uses it at each communication
    round or epoch to evaluate candidate model performance.
    Four evaluation metrics were selected from available \acrshort{mtsad}
    metrics~\cite{sorbo2024navigating,NEURIPS2024_c3f3c690} to ensure
    application-agnostic performance assessment. Specifically, the
    threshold-based metrics are point-wise F1-score and
    composite F1-score~\cite{9525836}, while the non-threshold-based metrics are
    \gls{aucpr} and \gls{vuspr} introduced by~\cite{10.14778/3551793.3551830}.

    \subsection{Datasets}\label{subsec:datasets}

    In this work, we use \gls{asd} proposed by~\cite{10.1145/3447548.3467075} as a benchmark that
    comprises 12 datasets designed for multivariate time series anomaly detection and interpretation.
    They are collected from a large internet company, each of the entities has 19 metrics characterizing status of the server.
    The time series constitutes 19 variables and data points that are equally spaced for every 5 minutes.
    The overall anomaly ratio is $4.61\%$.
    \acrshort{asd} is chosen for avoiding the following challenges: (i) Information Sparsity or Loss, (ii) Concept Drift,
    (iii) Mixed-valued Variables.

    In this paper, we propose~\gls{qappd}, a new collection of publicly available multivariate time series data~\cite{nosrati2026dataset}.
    It is explicitly made for \acrshort{mtsad} in federated learning. \acrshort{qappd} serves as the
    second dataset used for experimental evaluation in this study.
    It consists of ten independent train–test pairs, each of which can be interpreted as data local to
    federated learning edge nodes. It offers a physically grounded and semantically interpretable data-generation process.
    The dynamics were inspired by robotic pick and place operations, typically comprise multiple mechanical
    components---rotary joints, arms, and a gripper (mechanical hand) to pick and place items,
    carousels to transport these items, and visual inspection for verifying the existence the items and communicating their position.
    Along with all the structural components mentioned, the design considered nuances such as elevation of the platforms,
    composition of the components, distances to one another, and the proportions of these parts.
    To control the degree of statistical heterogeneity across datasets, the number of system components and
    the number of trajectory waypoints were kept constant across all ten train–test pairs,
    i.e.\ each configuration contains one articulated robot and two carousel.
    The resulting system behavior is characterized by cyclic trajectories in the system’s state space, where the gripper
    follows a smooth, periodic motion pattern resembling a cycloid-like motion.
    The trajectories were generated using the tool provided by~\cite{SSRH25}.

    To instantiate the above process, a cyber-physical system (\acrshort{cps}) called \allowbreak{Quan-ser} Aero 2 was used, i.e.\ a fully
    integrated dual-motor experimental setup, designed for advanced control research for aerospace and aeronautics applications.
    This device has two degrees of freedom (\acrshort{dof}), which are used to model a simplified robotic arm in this work.
    In this abstraction, the front fan is treated as the robotic gripper, while the body of the device represents a
    planar 2-\acrshort{dof} arm.
    The dataset variables are listed in Table~\ref{tab:table_qappd}.

    \paragraph{Dataset Design Principles.}
    The dataset was constructed using design semantics that explicitly encode normal
    and anomalous behavior. This approach significantly reduces the manual labeling effort, as anomalies are known by design.

    \paragraph{Properties of the Dataset.}
    The multivariate time series represents the system trajectory, with the observed
    variables jointly defining the system state at each discrete time step.
    In addition, the time series exhibits approximately periodic behavior.

    \begin{table}
        \centering
        \caption{Detailed CPS data on key observed variables.}
        \begin{tabular}{l@{\hspace{30pt}}l@{\hspace{30pt}}l@{\hspace{30pt}}l@{\hspace{30pt}}l}
              \toprule
            \textbf{Name} & \textbf{Description} & \textbf{Unit} & \textbf{Range} \\
            \midrule
              voltage0 & DC-motor 0 voltage & \si{\volt}  & $\pm$ 18\\
              voltage1 & DC-motor 1 voltage & \si{\volt} & $\pm$ 18\\
              current0 & DC-motor 0 current & \si{\ampere} & 0.540\\
              current1 & DC-motor 1 current & \si{\ampere} & 0.540\\
              motorSpd0 & Rotational speed of fan & rpm & 3050\\
              motorSpd1 & Rotational speed of fan & rpm & 3050\\
              pitch & Actual pitch angle & \si{\radian} & $\pm 2\pi/9$\\
              yaw & Actual yaw angle & \si{\radian} & $2\pi$\\
              pitchDot & Pitch's angular velocity & \si{\radian\per\second} & ---\\
              yawDot & Yaw's angular velocity & \si{\radian\per\second} & ---\\
            \bottomrule
        \end{tabular}\label{tab:table_qappd}
     \end{table}

    \paragraph{Anomaly Design from Two Sources.}

     \begin{itemize}
         \item Anomalies induced by system dynamics:
         \begin{itemize}
             \item Let $D=[0.3, 0.7]$ denote the set representing the  expected range of pause or delay duration
                   (in seconds) at pick-up anticipation positions. Then, any value falling outside $D$ is deemed anomalous.

             \item Let $P=[3,5]$ be the set representing normal pick-up angular
                   positions (in degrees). We consider any value not in $P$ anomalous.
         \end{itemize}

     \item Anomalies induced by the system controller:
         \begin{itemize}
            \item Let $\bar V=[-18,18]$ be the set representing the nominal voltage range for DC motors.
                  Then any value outside $\bar V$ is considered anomalous.
         \end{itemize}
     \end{itemize}

\begin{figure}
  \centering
  \includegraphics[width=\textwidth]{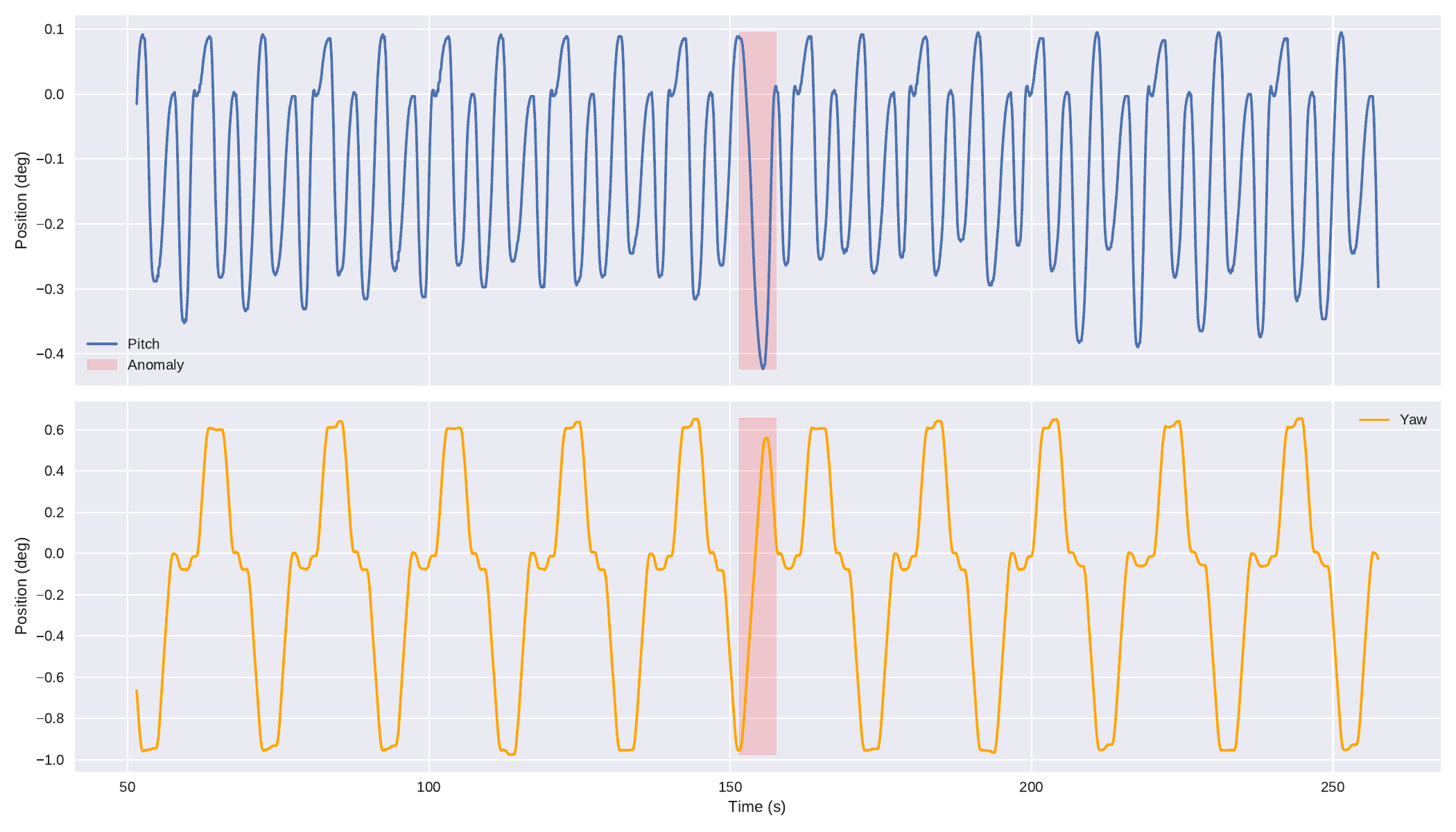}
  \caption{This snippet captures around 200 seconds of recorded data.
    Red highlights a misposition anomaly at a picking position affecting both yaw and pitch variables.}
  \label{fig:n_vs_a}
\end{figure}

    \paragraph{Data Acquisition.}
    The dataset was originally recorded at 500\,Hz and downsampled to 50\,Hz to reduce
    computational overhead, resulting in equally distant time steps of 0.02 seconds.
    \Cref{fig:n_vs_a} illustrates an example anomaly in the test set.

    \subsection{Baselines and Architectures}
    Five representative \gls{mtsad} baseline methods are selected based on their compatibility with the employed
    \acrshort{fl} software frameworks and their architectural suitability. In particular, methods requiring a single
    optimization phase are preferred, excluding approaches with additional optimization during evaluation (e.g., MAD-GAN~\cite{li2019mad}).
    The selection further considers inference latency via forward-pass time complexity, problem scope,
    and feature extraction capabilities, specifically the ability to model temporal (intra-variable) and spatial
    (inter-variable) dependencies. The curated methods and their characteristics are summarized in~\Cref{tab:tab_methods}.

     \begin{table}[H]
         \centering
         \caption{Summarizing how suitable the \acrshort{mtsad} methods are in terms of the aforementioned factors.
             Time complexity is per dominant layer(s) rather than total. $m$: sequence length. $d$: variable dimension.
             $z$: latent dimension. $k$: kernel size. $d=1$ for 1D convolution but for \acrshort{mtsad}, equals to the number
             of variables $m$. UTSAD: Univariate Time Series Anomaly Detection.}\label{tab:tab_methods}
         \begin{tabular}{l l c cc }
             \toprule
             \textbf{Method} & \textbf{Complexity} & \textbf{Problem} & \textbf{Temporal} & \textbf{Spatial} \\
             \midrule
             USAD~\cite{10.1145/3394486.3403392} & $O(m\cdot d\cdot z)$ & MTSAD & \xmark & \xmark \\
             DeepAnT~\cite{8581424} & $O(m\cdot d\cdot k\cdot z)$ & UTSAD & \cmark & \xmark \\
             LSTM-AE~\cite{8404689} & $O\big(m \cdot d\cdot z^2)$ & MTSAD & \cmark & \xmark \\
             TranAD~\cite{10.14778/3514061.3514067} & $O(m^2\cdot d\cdot z)$ & MTSAD & \cmark & \cmark \\
             MTAD-GAT~\cite{zhao2020multivariate} & $O(m^2\cdot d\cdot z + m\cdot d^2\cdot z)$ & MTSAD & \cmark & \cmark \\
         \bottomrule
         \end{tabular}
     \end{table}

    \section{Experimental Setup}\label{sec:setup}

    \paragraph{Software.} The computer cluster used to run experiments of this study, runs on Ubuntu
    20.04 LTS (Focal Fossa) and is managed for runs by SLURM Workload Manager.
    Python programming language was chosen for its versatility and the core implementations
    of this paper are done with it. Lightning is a high level framework for PyTorch,
    was selected to keep the unnecessary engineering code abstracted away, allowing us to focus on the problem-solving.
    Flower is chosen for \acrshort{fl} simulations, and a minimal framework is
    self-developed for \acrshort{hfl} simulations utilizing GPU for parallel processing.

    \paragraph{Hyperparameter Tuning.}
    To ensure fair comparisons, the algorithms are configured using their best searched hyperparameters preceding the
    main experiment runs.
    70\% of test and 80\% of train sets were sampled to tune the hyperparameters, speeding up the
    hyperparameter search as well as ensuring the representativeness of the sets. The objective for the search is based
    on maximizing point-wise F1-score.
    The batch size is set to 1024, however, for the trainings that are run with distributed training strategy, the
    batch size was divided by the number of GPUs to obtain the same results as if they are run on single GPU.
    The hyperparameter sweep were carried out using GPU's Tensor cores that support reduced precision format
    TensorFloat-32 (\textbf{TF32}), while using CUDA cores that support standard single-precision floating-point
    (\textbf{FP32}) for the main experiments.
    TF32 allows for using the same dynamic range as FP32 (8-bit exponent) but lower precision, i.e. 10-bit mantissa,
    compared to 23 bits.
    This is done to trade off some numerical accuracy with speed and memory for the heavy task of hyperparameter search.

    \paragraph{Hardware.}
    The experiments are performed on a single computer cluster node that includes 4x NVIDIA RTX A6000 graphic card with 48GB
    vRAM each, 2x AMD EPYC 7452 with 2.35GHz frequency with total of 64 computing cores.

    \paragraph{Main Experiments.}
    To obtain meaningful results, the experiments\footnote{The code for the experiments are publicly available \url{https://github.com/JRC-ISIA/industrial-federated-learning/}.} are independently run five times with different initialization seeds.
    The corresponding results for each unique experiment is reported as the mean plus-minus one standard deviation.
    The results are reported from the best fit state of the models.

    \section{Results}\label{sec:results}

    Detection performances are reported in~\Cref{tab:cl_fl_hfl_perf}. Under centralized learning, USAD, LSTM-AE,
    and DeepAnT outperform TranAD and MTAD-GAT overall. LSTM-AE shows stable performance across both datasets, USAD
    performs best on \acrshort{qappd}, and DeepAnT on \acrshort{asd}, despite its weaker performance on \acrshort{qappd}.
    MTAD-GAT and TranAD generally struggle in this setting. In federated learning, USAD and LSTM-AE achieve satisfactory
    results on \acrshort{qappd} but remain less competitive on \acrshort{asd}, whereas DeepAnT and MTAD-GAT perform better
    on \acrshort{asd}. TranAD is consistently the weakest method. In hierarchical federated learning, USAD and
    LSTM-AE perform best on \acrshort{qappd}, MTAD-GAT and DeepAnT excel on \acrshort{asd}, and TrandAD falls short on
    \acrshort{qappd} while moderately succeeds on \acrshort{asd}.

    \begin{table}[H]
        \caption{Detection performance across different setups, reported as
                 mean $\pm$ standard deviation over five independent runs.
                 F1 denotes the standard point-wise F1 score;
                 F1\textsubscript{c} denotes the composite F1 score.}
         \label{tab:cl_fl_hfl_perf}
         \centering
         \begin{tabular}{l l l c ccc}
              \toprule
              & & \textbf{Method} & \textbf{F1\textuparrow} & \textbf{F1\textsubscript{c}\textuparrow} & \textbf{VUS-PR\textuparrow} & \textbf{AUC-PR\textuparrow} \\
             \midrule
             \multirow{10}{*}{\rotatebox{90}{Centralized}}
              & \multirow{5}{*}{\rotatebox{90}{ASD}}
              & USAD & \underline{57.247 \std{0.122}}&61.369 \std{0.082} &45.676 \std{3.430} &42.020 \std{2.712} \\
              &  & DeepAnT &57.025 \std{0.000} &\textbf{84.166 \std{0.459}} &\underline{55.976 \std{0.725}} &\underline{51.147 \std{0.656}} \\
              &  & LSTM-AE &\textbf{63.456 \std{0.963}} &\underline{82.117 \std{3.394}} &\textbf{72.900 \std{2.027}} &\textbf{72.615 \std{1.768}} \\
              &  & TranAD &57.032 \std{0.009} &60.108 \std{2.362} &44.077 \std{0.493} &40.110 \std{0.129} \\
              &  & MTAD-GAT &57.033 \std{0.011} &66.179 \std{6.240} &45.942 \std{3.587} &41.216 \std{2.801} \\
             \cmidrule(l){2-7}
              & \multirow{5}{*}{\rotatebox{90}{QAPPD}}
              & USAD      &\textbf{23.849 \std{0.891}} &\underline{37.035 \std{2.728}}&\underline{13.030 \std{0.494}} &\underline{19.777 \std{0.768}} \\
              &  & DeepAnT  &11.711 \std{2.849} &18.338 \std{4.439} &6.613 \std{0.885} &6.110 \std{0.474} \\
              &  & LSTM-AE  &\underline{28.293 \std{8.387}} &\textbf{50.181 \std{10.358}}&\textbf{17.238 \std{5.619}} &\textbf{23.365 \std{6.600}} \\
              &  & TranAD   &13.910 \std{2.233} &18.706 \std{4.938} &7.907 \std{1.056} &7.159 \std{1.369} \\
              &  & MTAD-GAT &13.284 \std{1.547} &19.172 \std{4.908} &8.082 \std{0.591} &7.425 \std{1.349} \\
             \midrule
             \multirow{10}{*}{\rotatebox{90}{Federated}}
              & \multirow{5}{*}{\rotatebox{90}{ASD}}
              & USAD & 57.025 \std{0.000}& 57.398 \std{0.005} &43.325 \std{0.007} &41.044 \std{0.004} \\
              &  & DeepAnT &57.025 \std{0.000} &\textbf{85.382 \std{0.874}} &\textbf{55.379 \std{0.318}} &\textbf{50.664 \std{0.321}} \\
              &  & LSTM-AE &57.025 \std{0.000} &57.2785 \std{0.162} &43.544 \std{0.327} &41.407 \std{0.475} \\
              &  & TranAD &\textbf{57.062 \std{0.034}} &59.003 \std{1.897} &44.2354 \std{0.381} &40.076 \std{0.285} \\
              &  & MTAD-GAT &\underline{57.061 \std{0.039}} &\underline{62.553 \std{1.935}} &\underline{46.442 \std{1.183}} &\underline{41.859 \std{1.186}} \\
             \cmidrule(l){2-7}
              & \multirow{5}{*}{\rotatebox{90}{QAPPD}}
              & USAD      &\textbf{33.230 \std{0.155}} &\textbf{56.412 \std{0.398}}&\textbf{19.137 \std{0.109}} &\textbf{26.651 \std{0.054}} \\
              &  & DeepAnT  &13.0456 \std{0.375} &15.421 \std{0.586} &7.735 \std{0.076} &5.661 \std{0.157} \\
              &  & LSTM-AE  &\underline{25.139 \std{7.059}} &\underline{35.909 \std{14.888}}&\underline{13.872 \std{4.375}} &\underline{18.773 \std{5.920}} \\
              &  & TranAD   &12.486 \std{1.667} &16.195 \std{3.115} &6.503 \std{1.047} &5.905 \std{0.867} \\
              &  & MTAD-GAT &12.097 \std{1.110} &19.671 \std{3.481} &7.221 \std{0.424} &7.009 \std{0.715} \\
             \midrule
             \multirow{10}{*}{\rotatebox{90}{Hierarchical Federated}}
              & \multirow{5}{*}{\rotatebox{90}{ASD}}
              & USAD & 57.023 \std{0.000}&57.408 \std{0.005} &40.975 \std{0.006} &41.060 \std{0.005} \\
              &  & DeepAnT &\underline{57.024 \std{0.003}} &\underline{64.718 \std{2.681}} &\textbf{41.860 \std{0.310}} &41.039 \std{0.489} \\
              &  & LSTM-AE &57.023 \std{0.000} &57.189 \std{0.119} &41.340 \std{0.479} &\underline{41.295 \std{0.577}} \\
              &  & TranAD &\textbf{57.024 \std{0.002}} &60.423 \std{1.501} &41.323 \std{0.208} &40.079 \std{0.237} \\
              &  & MTAD-GAT & 57.023 \std{0.008} &\textbf{78.937 \std{2.255}} &\textbf{44.705 \std{0.608}} &\textbf{43.686 \std{0.528}} \\
             \cmidrule(l){2-7}
              & \multirow{5}{*}{\rotatebox{90}{QAPPD}}
              & USAD      &\textbf{34.612 \std{0.058}} &\underline{65.767 \std{0.141}}&\textbf{22.564 \std{0.024}} &\textbf{30.053 \std{0.045}} \\
              &  & DeepAnT  &11.631 \std{0.252} &13.468 \std{0.444} &5.275 \std{0.194} &5.212 \std{0.298} \\
              &  & LSTM-AE  &\underline{26.337 \std{7.008}} &\underline{47.693 \std{10.960}}&\underline{15.710 \std{4.754}} &\underline{21.476 \std{5.929}} \\
              &  & TranAD   &12.022 \std{2.624} &18.982 \std{7.489} &6.236 \std{0.967} &6.120 \std{1.483} \\
              &  & MTAD-GAT &11.855 \std{1.322} &33.084 \std{2.783} &7.356 \std{0.620} &8.820 \std{0.764} \\
             \bottomrule
         \end{tabular}
     \end{table}

\begin{figure}
  \centering
  \includegraphics[width=\textwidth]{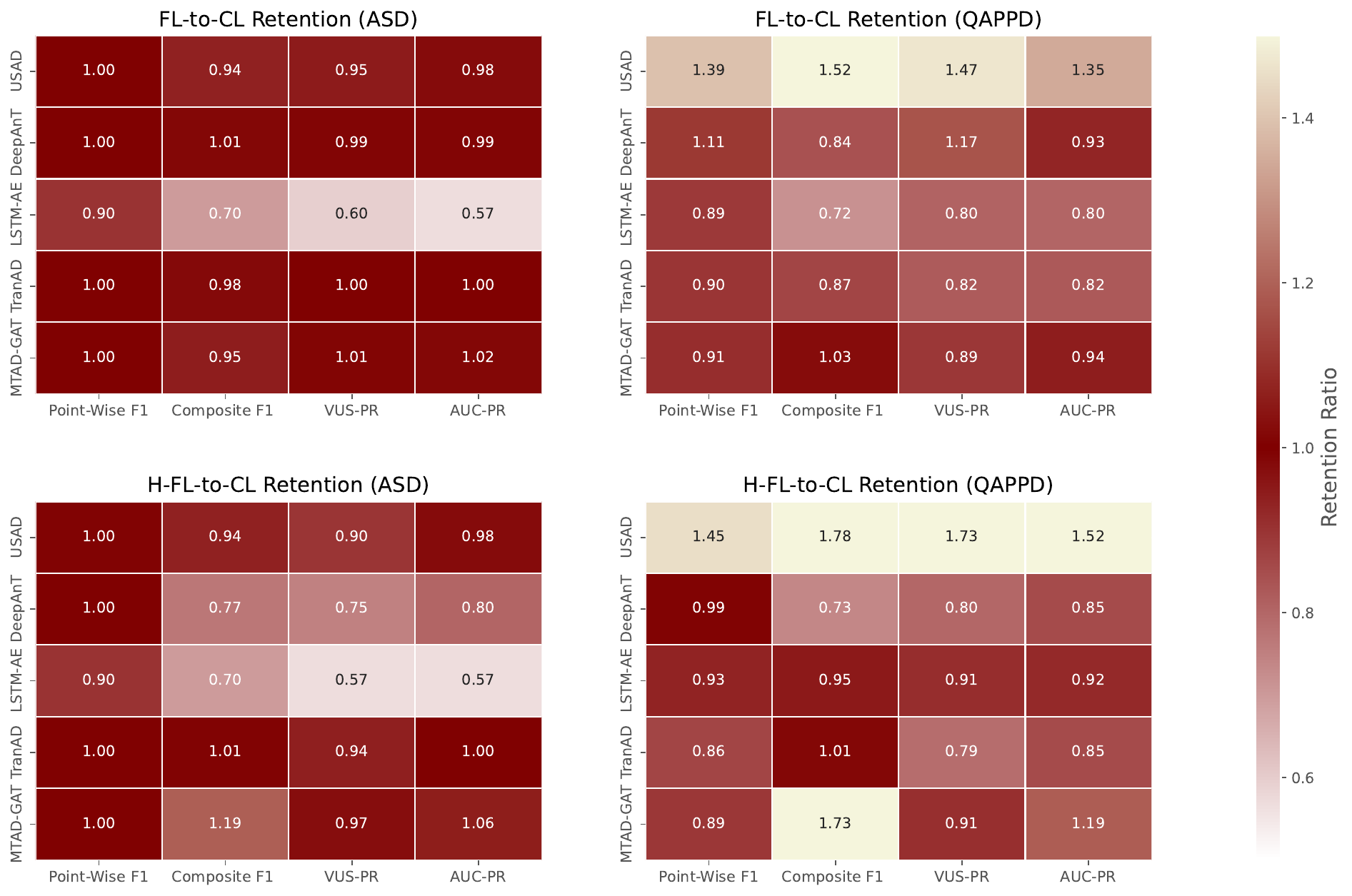}
  \caption{Relative performance with respect to \acrshort{cl}, where a ratio of 1.0 denotes parity.}
  \label{fig:p_g}
\end{figure}

    To quantify performance retention across the two datasets and three learning paradigms, the values of corresponding metrics
    in~\acrshort{fl} and~\acrshort{hfl} were divided by those of~\acrshort{cl} and shown in~\Cref{fig:p_g}.
    Each cell shows the ratio of a given metric value obtained under~\acrshort{fl} or~\acrshort{hfl} to that of~\acrshort{cl},
    illustrating the sensitivity of each method to the distributed learning paradigm and dataset characteristics.

    Notably, federated and hierarchical federated settings, although with different degrees of distribution,
    did not substantially affect model behaviors.
    MTAD‑GAT and TranAD were the most robust, showing the smallest performance drop.
    LSTM‑AE and DeepAnT suffered the largest degradations across metrics.
    Generally, USAD showed an unexpected improvement compared to other baselines.

    \section{Discussion}\label{sec:discussion}

    The empirical results indicate that architectural heterogeneity and the adoption of state-of-the-art models, such as
    Transformers and GNNs, do not inherently lead to improved anomaly detection performance. This suggests that increased
    model capacity and architectural sophistication may introduce unnecessary modeling overhead, particularly in repetitive
    industrial time series. Instead, less complex methods consistently achieve stronger results across settings.
    Notably, models that either explicitly capture temporal
    dependencies---such as LSTM- and TCN-based architectures---or employ effective training objectives, as in USAD with fully
    connected layers under an adversarial loss formulation, demonstrate superior performance. This suggests that alignment
    between model inductive bias, training strategy, and the characteristics of repetitive industrial time series is more
    critical than architectural complexity alone.

    \section{Conclusion}

    The comparative evaluation conducted in this paper reveals that the effectiveness of anomaly detection methods
    is tightly coupled to both the learning paradigm in which they operate and the characteristics of the datasets on
    which they are evaluated. These findings underscore the importance of considering learning topology and data structure
    jointly when assessing multivariate time series anomaly detection methods.
    A limitation of this work is that, while a new dataset is proposed and empirical benchmarking is provided,
    these results are not supported by a rigorous analysis that would enable deeper insights.
    Looking forward, several research directions naturally emerge from this work, including extension of anomaly definitions
    beyond bounded deviations to incorporate multiscale notions of abnormality, and integrating domain knowledge into
    learning objectives and selecting the architectural priors that align with the underlying system dynamics to support
    explainable and trustworthy anomaly detection.

    \begin{credits}
        \subsubsection{\ackname} The financial support by the Christian Doppler Research Association, the
        Austrian Federal Ministry for Digital and Economic Affairs and the Federal State
        of Salzburg is gratefully acknowledged.
    \end{credits}
%
%
    \bibliographystyle{splncs04}
    \bibliography{mtsad-references}

    \newpage
    \appendix
    \begin{center}
        {\LARGE\bfseries Appendix\par}
    \end{center}
    \vspace{1em}
    \section{Detail on Anomalies}\label{sec:anomaly_detail}

    Anomalies are categorized as $P$ (position), $D$ (delay), and $\bar V$ (voltage), as described in~\Cref{subsec:datasets}.
    \Cref{tab:tab_qappd} summarizes their statistics.

         \begin{table}
         \centering
         \caption{Statistics of anomalies in \acrshort{qappd}. The voltage anomaly appears as a long contiguous subsequence
         only in a single dataset, ensuring a comparable overall
         anomaly proportion with the shorter, more frequent position and delay anomalies.
            }\label{tab:tab_qappd}
         \begin{tabular}{p{0.2\textwidth} p{0.07\textwidth} p{0.07\textwidth} p{0.07\textwidth} p{0.07\textwidth} p{0.07\textwidth} p{0.07\textwidth} p{0.07\textwidth} p{0.07\textwidth} p{0.07\textwidth} p{0.07\textwidth} }
             \toprule
             ID & 1 & 2 & 3 & 4 & 5 & 6 & 7 & 8 & 9 & 10 \\
             \midrule
             Type & $P$,$D$ & $D$ & $P$ & $P$ & $D$ & $P$,$D$ & $\bar V$ & $D$ & $P$,$D$ & $P$,$D$ \\
             Frequency & 2.3\% & 1.3\% & 2.1\% & 2.7\% & 1.0\% & 1.0\% & 10.1\% & 1.0\% & 2.0\%  & 2.3\% \\
             Count & 4 & 3 & 4 & 5 & 3 & 2 & 1 & 3 & 4 & 4 \\
         \bottomrule
         \end{tabular}
     \end{table}

\end{document}